\documentclass[10pt,twocolumn,letterpaper]{article}

\usepackage{cvpr}              
\usepackage{url}
\usepackage[table,xcdraw]{xcolor}
\usepackage{colortbl}
\usepackage{booktabs}
\usepackage{graphicx}
\usepackage{subcaption}
\usepackage{svg}
\usepackage{amsmath}
\usepackage{bm}
\usepackage{wrapfig}
\usepackage{color}
\usepackage{enumerate}
\usepackage{siunitx}
\usepackage{bbding}
\usepackage{multirow}
\usepackage{multicol}
\definecolor{mineblue}{rgb}{0.929,0.949,0.988}
\usepackage[most,skins,theorems]{tcolorbox}

\tcbset{
  aibox/.style={
    width=\linewidth,
    top=4pt,
    bottom=4pt,
    colback=mineblue,
    colframe=black,
    colbacktitle=black,
    enhanced,
    center,
    attach boxed title to top left={yshift=-0.1in,xshift=0.15in},
    boxed title style={boxrule=0pt,colframe=white,},
  }
}
\usepackage[most]{tcolorbox}
\usepackage{listings}
\usepackage{etoolbox}

\lstdefinestyle{prompttxt}{
  basicstyle=\ttfamily\small,
  columns=fullflexible,
  breaklines=true,
  breakatwhitespace=false,
  keepspaces=true,
  showstringspaces=false,
  tabsize=2
}

\newtcolorbox{promptbox}[1]{
  enhanced,
  breakable,
  colback=gray!1,                
  colframe=black!40,             
  boxrule=0.6pt,
  arc=2mm,
  outer arc=2mm,
  left=6pt, right=6pt, top=6pt, bottom=6pt,
  title={#1},
  coltitle=white,                
  fonttitle=\bfseries,
  colbacktitle=black!80,         
  boxed title style={arc=1mm, outer arc=1mm},
  attach boxed title to top left={yshift=-2mm, xshift=2mm},
  drop shadow                    
}

\newcommand{\promptboxfromfile}[2]{%
  \begin{promptbox}{#1}
    \lstinputlisting[style=prompttxt]{#2}
  \end{promptbox}%
}

\newtcolorbox{AIbox}[2][]{aibox,title=#2,#1}

\newcommand{\scinot}[2]{#1\times 10^{#2}}
\definecolor{cvprblue}{rgb}{0.21,0.49,0.74}
\usepackage[pagebackref,breaklinks,colorlinks,allcolors=cvprblue]{hyperref}

\def\paperID{15825} 
\def\confName{CVPR}
\def\confYear{2026}

\title{\includegraphics[height=1em]{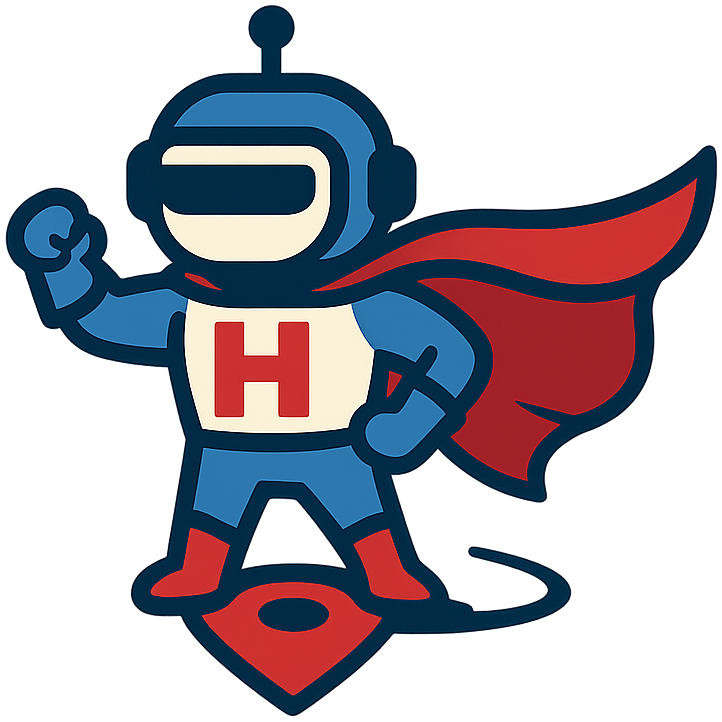} HiRO-Nav: \textbf{H}ybr\textbf{i}d \textbf{R}eas\textbf{O}ning Enables Efficient Embodied \textbf{Nav}igation}

\author{
He Zhao$^{1}$ \quad
Yijun Yang$^{2}$ \quad
Zichuan Lin$^{2}$ \quad
Deheng Ye$^{3}$ \quad
Chunyan Miao$^{1}$ \\
$^{1}$Nanyang Technological University \\
$^{2}$Tencent \\
$^{3}$Independent Researcher \\
\tt\small ZHAO0378@e.ntu.edu.sg
}

\begin{document}
\maketitle
\begin{abstract}
Embodied navigation agents built upon large reasoning models (LRMs) can handle complex, multimodal environmental input and perform grounded reasoning per step to improve sequential decision-making for long-horizon tasks.
However, a critical question remains: \textit{how can the reasoning capabilities of LRMs be harnessed intelligently and efficiently for long-horizon navigation tasks?} In simple scenes, agents are expected to act reflexively, while in complex ones they should engage in deliberate reasoning before acting.
To achieve this, we introduce \textbf{H}ybr\textbf{i}d \textbf{R}eas\textbf{O}ning \textbf{Nav}igation (\textbf{HiRO-Nav}) agent, the first kind of agent capable of adaptively determining whether to perform thinking at every step based on its own \textit{action entropy}. 
Specifically, by examining how the agent's action entropy evolves over the navigation trajectories, we observed that only a small fraction of actions exhibit high entropy, and these actions often steer the agent toward novel scenes or critical objects. Furthermore, studying the relationship between action entropy and task completion (i.e., Q-value) reveals that improving high-entropy actions contributes more positively to task success.  
Hence, we propose a tailored training pipeline comprising hybrid supervised fine-tuning as a cold start, followed by online reinforcement learning with the proposed hybrid reasoning strategy to explicitly activate reasoning only for high-entropy actions, significantly reducing computational overhead while improving decision quality.
Extensive experiments on the \textsc{CHORES}-$\mathbb{S}$ ObjectNav benchmark showcases that HiRO-Nav achieves a better trade-off between success rates and token efficiency than both dense-thinking and no-thinking baselines. 
\end{abstract}   
\section{Introduction}
\label{sec:intro}

\begin{figure}[h]
    \centering
    \includegraphics[width=1.0\linewidth]{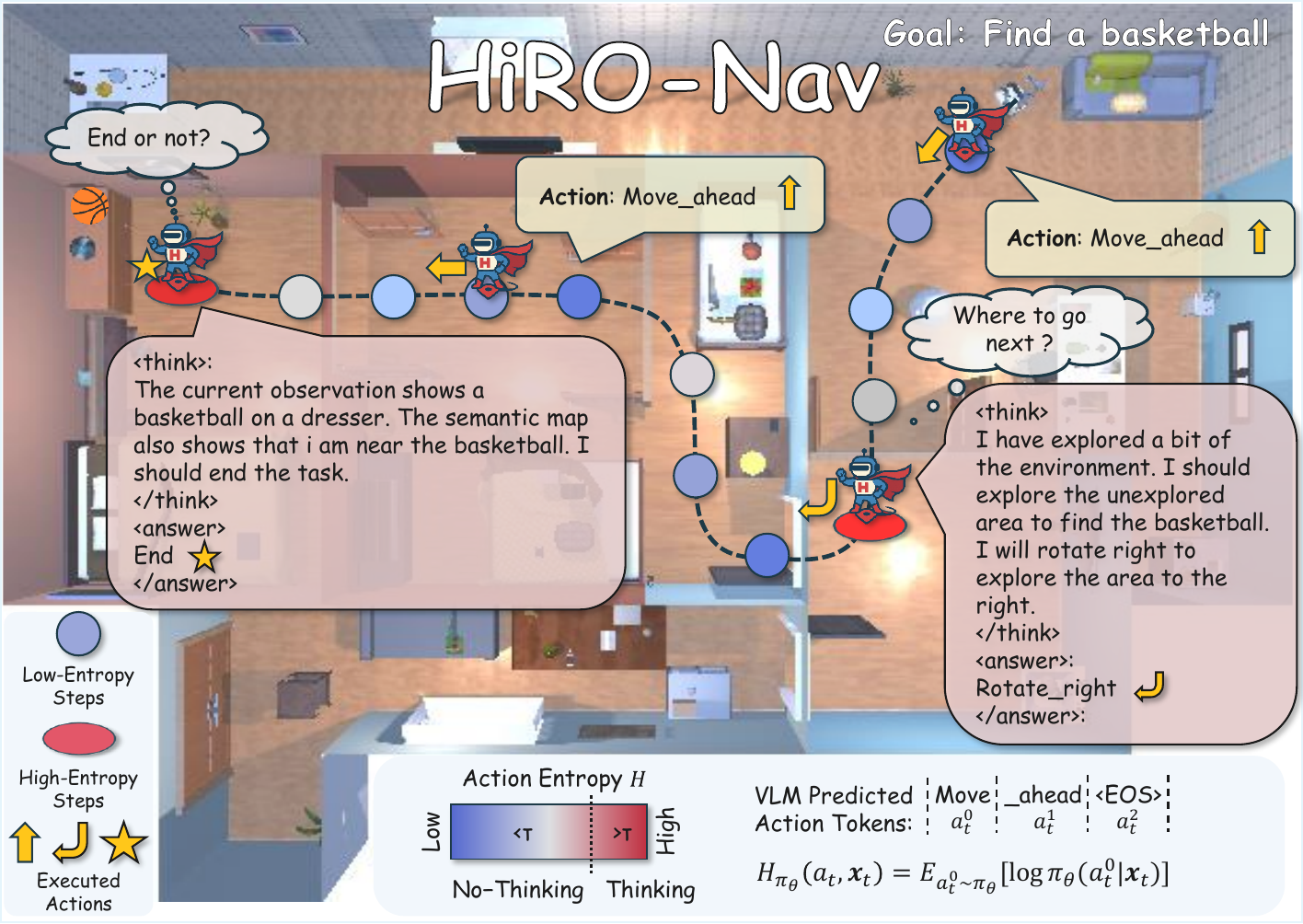}
    \caption{Illustration of HiRO-Nav agent adaptively determining whether to perform thinking based on its own action entropy $H_{\pi_\theta}(a_t,\bm x_t)$. Based on our observations, high-entropy actions often steer the agent toward novel scenes or critical objects, which are located on key waypoints over the navigation trajectory, i.e., the black dashed line in the bird-eye-view map. HiRO-Nav accordingly activates reasoning only for these high-entropy actions (red points), improving the trade-off between reasoning efficiency and performance compared to dense-thinking and no-thinking agents, as demonstrated in Fig.~\ref{fig:intro_tc}. \looseness-1}
     \label{fig:intro_overview}
\end{figure}

Embodied navigation aims to empower autonomous agents with the capability to perceive multimodal environmental information and to make decisions step by step for executing long-horizon tasks. Recent advancements in Large Reasoning Models (LRMs) such as DeepSeek-R1~\cite{deepseekr1}, Gemini2.5-Pro~\cite{gemini2.5pro} and o3~\cite{o3}, have demonstrated substantial improvements in perception and decision making ability by applying the Chain-of-Thought (CoT) \cite{cot} technique.
Leveraging these advancements, existing research efforts \cite{vln-r1, sgnav, cognav, mem2ego, cl-cotnav, apexnav} aim to develop embodied navigation systems grounded in LRMs. However, a critical question still remains:
\begin{AIbox}{}
\textbf{How can the reasoning capabilities of LRMs be harnessed intelligently and efficiently for long-horizon navigation tasks?}
\end{AIbox}

\begin{figure*}[t]
    \centering
    \begin{subfigure}[t]{0.26\textwidth}
    \centering
    \includegraphics[width=\textwidth]{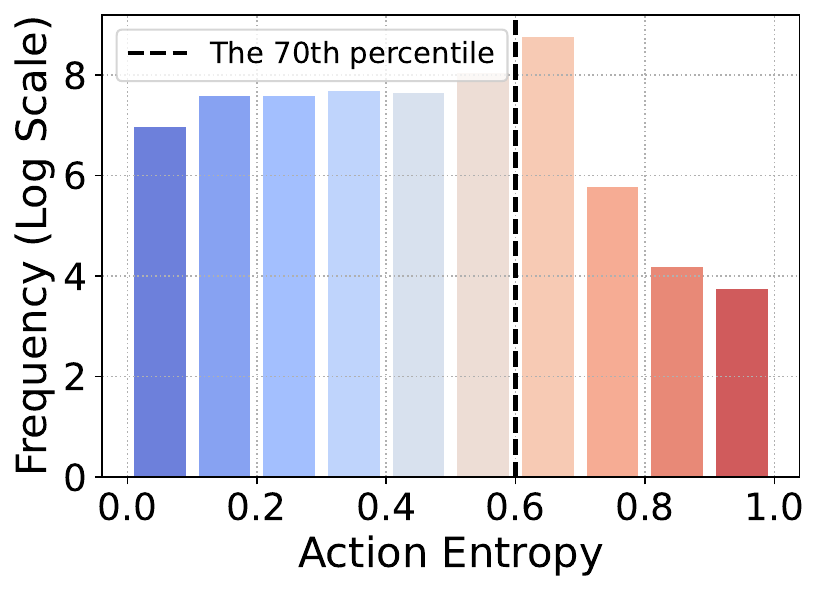}
    \caption{Distribution of AE.}
    \label{fig:entropy_dist}
  \end{subfigure}
  \quad
  \begin{subfigure}[t]{0.71\textwidth}
    \centering
    \includegraphics[width=\textwidth]{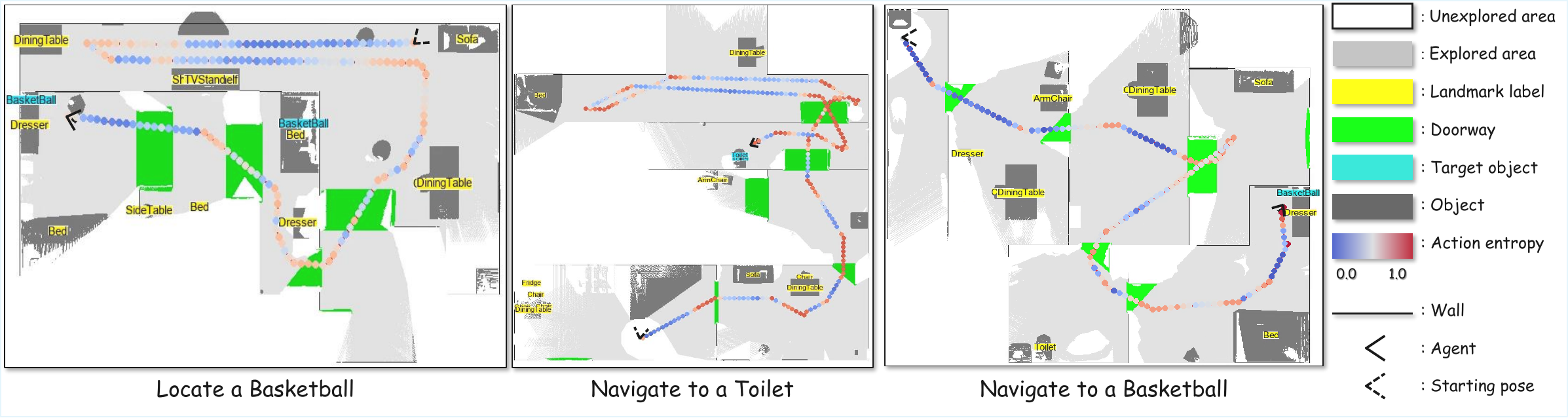}
    \caption{Visualization examples of AE at each waypoint on annotated semantic maps. \looseness-1}
     \label{fig:entropy_map}
  \end{subfigure}
  
\caption{\textbf{The distribution of action entropy (AE) over navigation trajectories}. We analyze the AE distribution of a VLM agent fine-tuned using expert trajectories on \textsc{CHORES}-$\mathbb{S}$ ObjectNav tasks. (a): Only a small fraction ($\sim$30\%) of actions exhibits high entropy (AE $\geq$ 0.6). (b): High-entropy actions (red points in the map) often steer the agent to explore novel areas or approach critical objects. An extended version of the figure (b) can be found in Fig.~\ref{fig:entropy_example_app} of Appendix.}
\label{figs:entropy_pattens}
\end{figure*}

\begin{figure}
    \centering
    \includegraphics[width=0.8\linewidth]{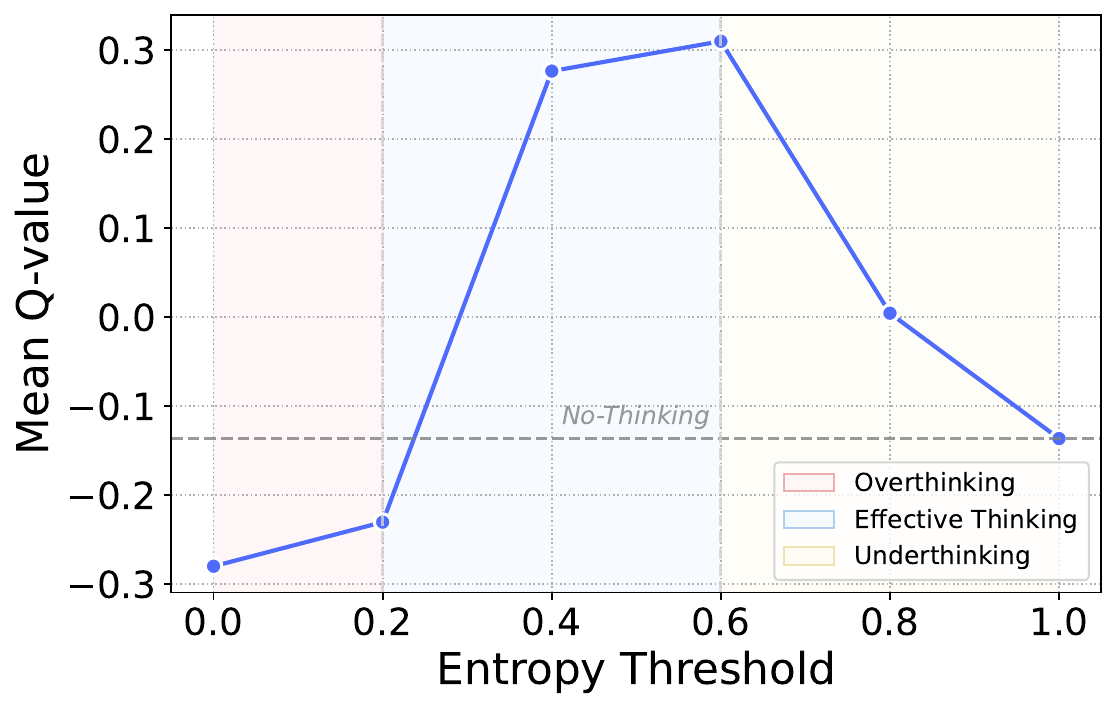}
    \caption{\textbf{Mean Q-value of a hybrid fine-tuned model introduced in Sec.~\ref{sec:sft} across various action entropy thresholds}. The high threshold means sparse activation of reasoning, resulting in high token efficiency. We conclude that thinking only for high-entropy actions (threshold$=$0.6) achieves the best trade-off between task completion and maximizing token efficiency. Lower or higher thresholds can result in ``overthinking'' or ``underthinking'' respectively, both degrading the final performance.\looseness-1}
    \label{fig:entropy-qvalue}
\end{figure}

Existing LRM-based navigation agents typically make decisions by thinking step by step \cite{cl-cotnav, apexnav, mem2ego, sgnav}. However, this dense thinking paradigm inevitably introduces substantial computational overhead, leading to increased latency that can impair efficiency in long-horizon and real-time navigation tasks. Furthermore, recent studies \cite{cot-or-not,think-or-not-think,when-to-think,overthinksurvey} reveal an \textit{Overthinking} phenomenon, where excessive reasoning in simple scenes can cause more hallucinations that diminish performance gains brought by the test-time scaling and eventually harm overall task completion. 
Therefore, agents are expected to act reflexively based on their perceptual capabilities. In contrast, when faced with complex scenes, such as encountering a crossroad leading to different unseen rooms, agents should carefully consider their options to select the most appropriate direction for future exploration. Unfortunately, LRM-based navigation agents with the ability to adaptively engage in reasoning remain underexplored.\looseness-1

Following recent literatures that revisit the entropy dynamics of LRMs \cite{80-20-rule, policy-entropy}, we performed an in-depth analysis on how the agent's action entropy (see Eq.~(\ref{eq:entropy}) for the formal definition) evolves across the ongoing navigation process.
As illustrated in Fig. \ref{figs:entropy_pattens}, we observed that only a small fraction of actions exhibit high-entropy, and these actions typically occur in complex scenes, steering the agent towards novel scenes or critical objects, such as leading the agent to explore a new room or take a novel item, whereas low-entropy actions are often taken in simple scenes, directing the agent to move straightforwardly from one location to another. 
Furthermore, we investigate the relationship between taking explicit reasoning  for actions with different levels of entropy and task completion (i.e., Q-value \cite{q-learning}). To this end, we restrict the agent to activate thinking for actions with entropy exceeding a specified threshold, and calculate the mean Q-value of executed actions across all navigation processes. As shown in Fig.~\ref{fig:entropy-qvalue}, we found that by thinking only for high-entropy actions, the LRM-based agent successfully mitigates overthinking and achieves the best trade-off between improving task completion and minimizing reasoning effort, demonstrating that improving high-entropy actions contributes more positively to task success. 
\begin{figure}[t]
    \centering
    \includegraphics[width=0.8\linewidth]{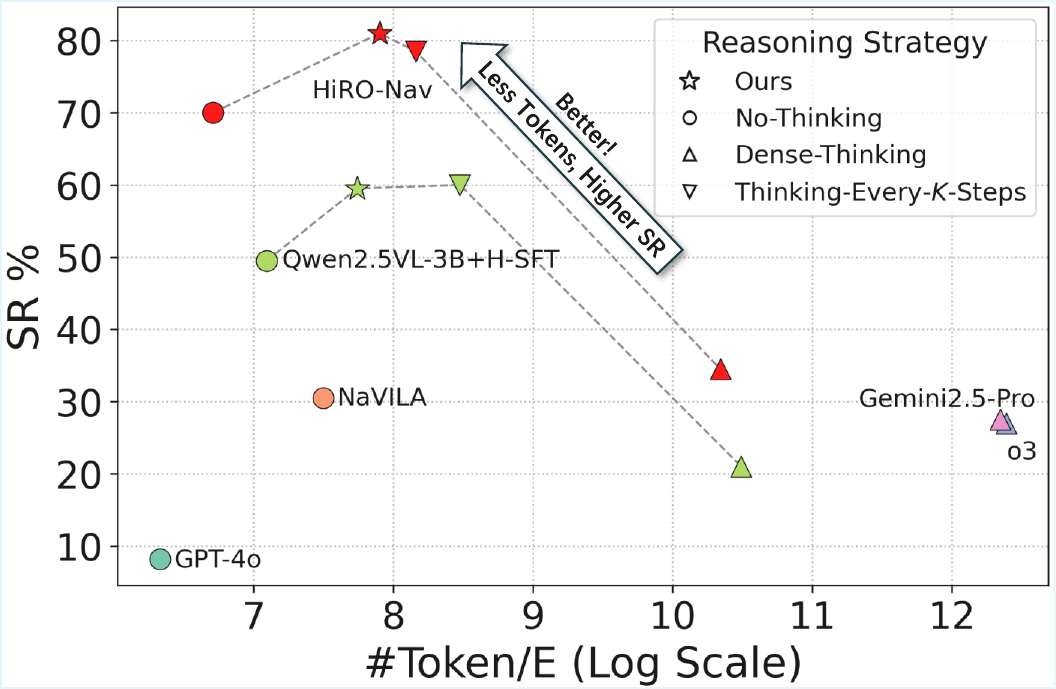}
    \caption{\textbf{Comparison of HiRO-Nav agent against SOTA baselines in terms of the trade-off between navigation success rate (SR) and token efficiency}. We compute the average number of model-generated tokens per episode (\#Token/E). HiRO-Nav with hybrid reasoning strategy (Ours) achieves the best trade-off.}
    \label{fig:intro_tc}
\end{figure}

Based on the aforementioned findings, we propose \textbf{H}ybr\textbf{i}d \textbf{R}eas\textbf{O}ning \textbf{Nav}igation (\textbf{HiRO-Nav}) agent, the first end-to-end navigation agent with a novel hybrid reasoning strategy, where the agent divides action into two classes: high and low entropy, based on a predefined threshold, and activates reasoning only for high-entropy ones, as illustrated in Fig.~\ref{fig:intro_overview}. To achieve this, we design a training pipeline comprising hybrid supervised fine-tuning, followed by online reinforcement learning with a adaptive reasoning strategy. Specifically, we fine-tune a vision language model (VLM) as a cold start on a carefully curated hybrid reasoning dataset. During online RL training, we found that directly training hybrid-thinking ability raises a \textit{model collapse} problem, in which the agent's no-thinking ability decreases significantly, limiting the overall performance. Hence, we split the RL training into two stages to optimize the no-thinking and thinking abilities separately. In stage I, we collect rollouts in no-thinking mode and solely optimize the agent's no-thinking ability. In stage II, we collect rollouts with proposed hybrid reasoning strategy and solely optimize the agent's thinking ability while maintaining the no-thinking ability through a KL regularization. \looseness-1

Extensive experiments on the \textsc{CHORES}-$\mathbb{S}$ ObjectNav benchmark~\cite{spoc} verify that HiRO-Nav with hybrid reasoning achieves a state-of-the-art trade-off between task success rate and reasoning efficiency compared with existing reasoning strategies in navigation, as shown in Fig. \ref{fig:intro_tc}, demonstrating that our reasoning strategy successfully avoids the overthinking problem while efficiently incentivizing the LRM’s reasoning capability. Furthermore, Pass@k evaluation results demonstrate a strong upper bound of HiRO-Nav’s navigation capability, even when using noisy annotated semantic maps as long-term memory.

\section{Related Work}
\subsection{Hybrid Reasoning of LRMs in Navigation}
LRMs like Deepseek-R1 \cite{deepseekr1} have made promising achievements in complex reasoning tasks by using CoT \cite{cot}. Recent works incentivize models to adaptively adjust their reasoning length or switch between different reasoning modes \cite{when-to-think,adactrl,adaptthink,lhrms,wang2025adaptive} to solve the overthinking problem \cite{think-or-not-think,cot-or-not, overthinking,yue2025don}. 
However, when to think in navigation tasks still remains underexplored. Existing LRM-based navigation agents typically perform thinking step by step \cite{vln-r1, sgnav, cognav, mem2ego, apexnav, cl-cotnav} and only a few works explore the reasoning strategy. OctoNav \citep{octonav} performs CoT reasoning every k steps during testing. Aux-Think \cite{aux-think} proposes to sft the model with a mixture of reasoning data and action-only data, while outputting actions only when testing. However, these existing methods engage in reasoning by predefined rules regardless of navigation dynamics, which will hinder model performance eventually. To this end, we are the first to propose a navigation agent with hybrid reasoning ability adaptively determining whether to perform thinking to achieve better navigation performance and efficiency. Detailed discussion of related works can be found in Appendix \ref{sec:app_related_work}. \looseness-1


\section{Hybrid Reasoning Navigation Agent}

In this Sec., we introduce the design principle of HiRO-Nav, which identifies two primary challenges and proposes the corresponding solutions as follows:
\begin{itemize}
    \item \textbf{When should the agent engage in deliberate reasoning?} In Sec.~\ref{sec:critical_step}, we identify that reasoning is necessary only for high-entropy actions through an in-depth analysis of how action entropy evolves over ongoing navigation processes and propose a hybrid reasoning strategy accordingly. \looseness -1
    \item \textbf{How to train the agent capable of hybrid reasoning ability?} We curate a training pipeline (Fig.~\ref{fig:model}) comprising hybrid supervised fine-tuning as a cold start (Sec.~\ref{sec:sft}), followed by a two-stage online reinforcement learning with the proposed hybrid reasoning strategy (Sec.~\ref{sec:onlineRL}).\looseness-1
\end{itemize}

\subsection{Problem Formulation} 
\label{sec:problem_setup}
In this work, we focus on the Object Goal Navigation (ObjectNav) task \citep{objectnav}, which requires agents to locate the predefined target object category in novel environments. Each task can be formulated as a Partially Observable Markov Decision Process (POMDP), denoted as $(S, A, O, T, R)$, where: $S$ denotes the state space; $A$ is the action space in textual form, $O$ represents the observation space, $T$ denotes the state transition function $s_{t+1}\sim T(s_{t+1}|s_t, a_t)$, $R$ encapsulates the reward function $R:S\times A \rightarrow \mathbb{R}$. An agent serves as a policy $\pi_{\theta}(\bm{y_t}|I, o_{t-w:t}, a_{t-w:t-1}, m_t)$ by generating textual output $\bm y_t$ based on the natural language instruction $I$ and history information including a short-term memory window $w$ and a long-term memory $m_t$. If LRM generates in thinking mode, then $\bm y_t = (v_t, a_t)$, where $v_t$ is the CoT reasoning trace. Otherwise, $\bm y_t = a_t$.
Following prior work, we adopt annotated semantic map (ASM) \cite{mapnav} as the long-term memory. 
For simplicity, we use $\bm x_t = (I, o_{t-w:t}, a_{t-w:t-1}, m_t)$ to represent the agent's input. More details about the observation and action spaces can be found in Appendix.

\subsection{A Comprehensive Analysis of Action Entropy}
\label{sec:critical_step}

Inspired by prior work revisiting the entropy dynamics of LRMs \cite{80-20-rule, policy-entropy, arpo}, we analyzed how action entropy evolves across long-horizon navigation trajectories. First of all, the action entropy is defined as below.
\begin{equation}
    H_{\pi_\theta}(a_t,\bm x_t) = \mathbb{E}_{a_t^0 \sim\pi_\theta}[\log \pi_\theta(a_t^0| \bm x_t)]
    \label{eq:entropy}
\end{equation}
where $a_t^0$ is the first token\footnote{We use the first token entropy instead of the mean of full sequence for higher computing efficiency. Both methods exhibit the similar pattern and performance. Detailed analysis can be found in Appendix \ref{sec:action_entropy}} of the predicted action $\bm a_t$ at timestep $t$. We show the results on \textsc{CHORES}-$\mathbb{S}$ ObjectNav benchmark using a Qwen2.5VL-3B \cite{qwen25vl} model finetuned on the collected expert trajectories and observe entropy's distribution pattern as below.

\textbf{Only a small fraction of actions exhibit high entropy.} As illustrated in Fig.~\ref{fig:entropy_dist}, the overall action entropy exhibits a mildly long-tail distribution. We found that 30\% of actions exhibits high entropy greater than 0.6, with only a few exceeding 0.7. \looseness-1

\textbf{High-entropy actions often steer the agent toward novel scenes or critical objects.} We plotted action entropy with the heat bar at each waypoint of the annotated semantic map in Fig.~\ref{fig:entropy_map} and observed that high-entropy actions typically occur in complex scenes, steering the agent to explore novel areas or approach critical objects. For example, at turning waypoints, high-entropy actions guide the agent to collect unseen environmental information; when the agent is close to the target object, high-entropy actions assess whether the success criteria have been met and determine the termination of the task. In contrast, low-entropy actions occur in simple scenes, directing the agent to move straightforwardly from one location to another, serving as transitions between trivial waypoints. Intuitively, in complex scenes, the agent requires careful reasoning to effectively guide the navigation process, whereas in simple scenes, acting directly without additional reasoning is sufficient.

To verify this, we investigate the relationship between taking explicit reasoning  for actions with different levels of entropy and task completion (i.e., Q-value \cite{q-learning}).
The Q-value is defined as
\begin{align}
    Q_{\pi_{\theta}}(s, a) &= \mathbb{E}_{a_t \sim\pi_{\theta}}\left[\sum_{t=0}^{\infty}\gamma^{t}R(s_t, a_t)|s_0=s, a_0=a\right] \notag \\
    &\approx \frac{1}{N}\sum^{N}_{n}\sum_{t=0}^{T}\gamma^{t}R(s^n_t, a^n_t)         
\end{align}
where $N$ is the number of sampled trajectories and $T$ is the maximum episode length. We use the discounted return as the Monte Carlo estimation of the Q-value. Here we set $\gamma$ to the commonly used 0.99.  We analyze a hybrid fine-tuned model with both no-thinking and thinking modes, as introduced in Sec.~\ref{sec:sft}. If the action entropy exceeds a threshold, we prompt the model to activate thinking mode and regenerate an action with CoT reasoning. Higher thresholds indicate lower reasoning frequency, resulting a higher token efficiency.
The results in Fig.~\ref{fig:entropy-qvalue} show that that thinking only for high-entropy actions (threshold=0.6) achieves the best trade-off between task completion and maximizing token efficiency. Lower thresholds (0$\sim$0.2) causes the agent thinking for low-entropy actions, leading to overthinking, whereas higher thresholds (\textgreater 0.6) force the agent to think less, resulting in underthinking. Both cases lead to degradation in task completion, demonstrating that deliberate reasoning is necessary only for high-entropy actions. \looseness-1

\begin{figure*}[tp]
    \centering
    \includegraphics[width=\linewidth]{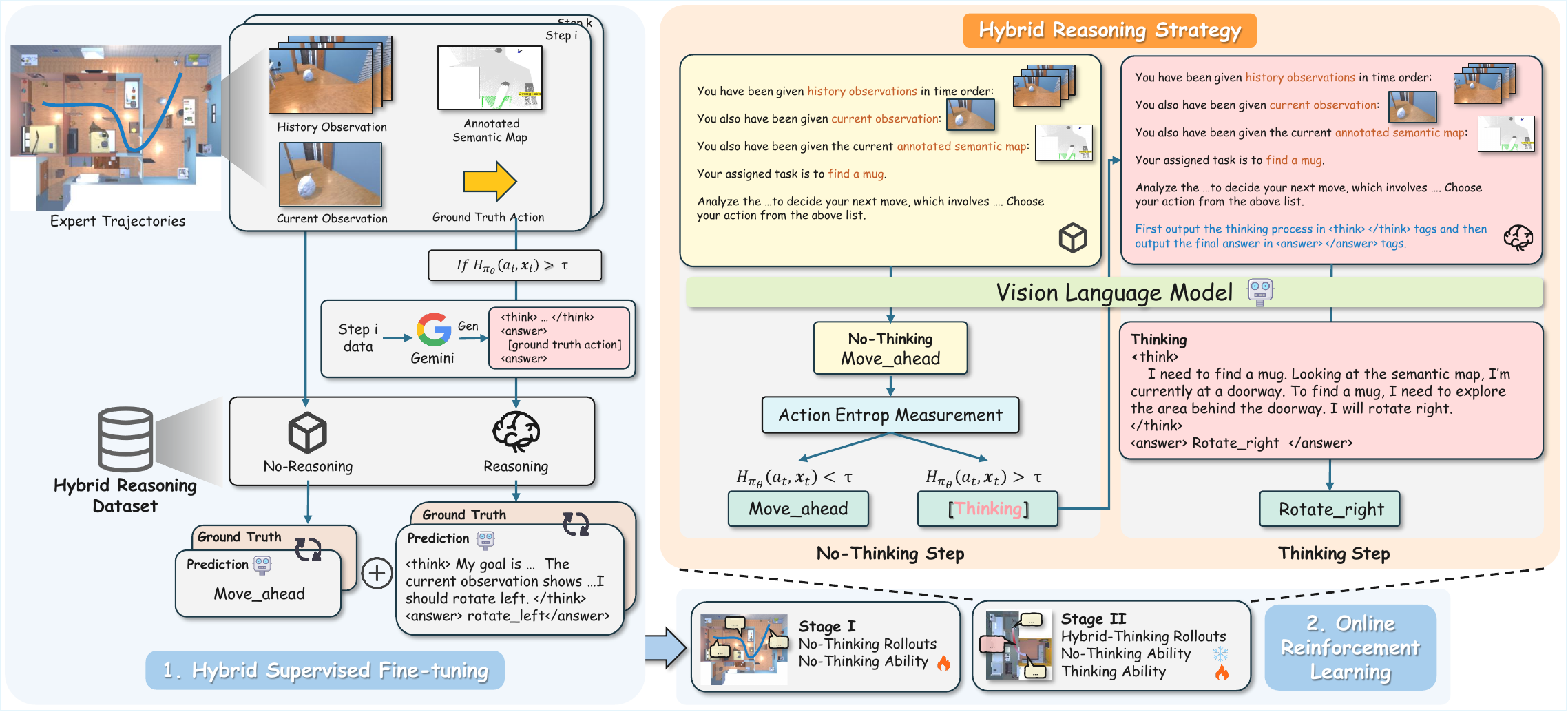}
    \caption{Overview of HiRO-Nav training pipeline and the proposed hybrid reasoning strategy. 
    which consists of two part: 
    (1) \textbf{Hybrid supervised fine-tuning} (left). We first collect a hybrid reasoning dataset (HRD) containing a no-reasoning dataset and a reasoning dataset annotated by Gemini2.0-Flash on high-entropy actions . 
    We then fine-tune a VLM on the HRD as a cold start to enable the agent with hybrid-thinking abilities. 
    (2) \textbf{Two-stage online reinforcement learning} (bottom right). In stage I, we collect no-thinking rollouts and train the agent's no-thinking ability. In Stage II, we collect rollouts with proposed hybrid reasoning strategy and train agent's thinking ability. 
    The proposed \textbf{Hybrid reasoning strategy} (top right) encourages the agent to activate thinking only for high-entropy actions.}
     \label{fig:model}
\end{figure*}

In summary, we design a novel \textbf{Hybrid Reasoning Strategy} that encourages the model to think only for high-entropy actions. As illustrated on the right part of Fig.~\ref{fig:model}, the agent initially predicts actions in no-thinking mode and then decides whether to activate thinking based on its action entropy. If it exceeds a threshold $\tau$, the model’s thinking mode is activated via prompting and then performs thinking before generating a new action. 
Moreover, we found that the hybrid reasoning strategy may encounter the repetition problem: the agent repeatedly predicts high-entropy actions, resulting in overthinking. To this end, we introduce a \textit{No-Thinking Window (NTW)}, which restricts the model to think every $K$ steps. \looseness-1

\subsection{Hybrid Supervised Fine-tuning}
\label{sec:sft}
As illustrated in the top left part of Fig.~\ref{fig:model}, an RL-trained policy \cite{poliformer} is adopted to collect expert trajectories based on its training environments, containing 2.86 million pairs of multi-modal observations and actions. In addition, we use the annotated semantic map (ASM) \cite{mapnav} as the long-term memory $m$ to compress the context of the agent.

To enable VLM agents to perform high-quality CoT reasoning before taking actions, we curate a reasoning dataset by selecting high-entropy action samples from the collected expert dataset. Specifically, a VLM is fine-tuned on the expert dataset and then used to select data according to the model-predicted action entropy. We only preserve top 20\% high-entropy samples. Next, we leverage the latest Gemini-2.0-flash model \citep{gemini} to generate reasoning traces for this data. Refer to Appendix \ref{sec:hrd_app} for more details. 
In summary, we collect a compact dataset containing 280K of high-quality reasoning-action pairs. 

The Qwen2.5-VL-3B is trained by \textbf{Hybrid Supervised Fine-Tuning} (H-SFT) using the curated reasoning dataset and no-reasoning dataset together, enabling the resulting agent to generate actions in two modes according to different prompts. Given the hybrid reasoning dataset $D_{\text{HRD}}=\{(\bm{x}_i, \bm{y}_i)\}_{i=1}^{|D|}$, the optimization objective is formulated as below. \looseness-1
\begin{equation}
    L_{\text{H-SFT}}(\theta) = -\mathbb{E}_{(\bm{x}, \bm{y})\sim D_{\text{HRD}}}\left[\sum_{l=1}^{|\bm{y}|}\log \pi_{\theta}(y_l|\bm x, y_{<l})\right] 
\end{equation}
where $y_l$ is the $l^{\text{th}}$ token of $\bm y$. 

\subsection{Online RL with the Hybrid Reasoning Strategy}
\label{sec:onlineRL}
Bulit top of on the SFT model, we use the PPO algorithm \cite{ppo,rlhf} to post-train it with the proposed hybrid reasoning strategy on the training tasks from \citet{poliformer}. However, a severe problem is observed if we directly train hybrid reasoning ability through RL: the agent's no-thinking performance exhibits a significant drop, eventually setting a ceiling on the success rate. We attribute this problem to the imbalanced training objective between reasoning and non-reasoning responses, in which the number of tokens generated in the thinking mode far exceeds that of the no-thinking mode, resulting in larger gradients derived by those thinking samples. \looseness-1


Hence, we propose a \textbf{Two-Stage Online Reinforcement Learning} strategy. In the first stage, we collect rollouts using the no-thinking prompt and then train the agent's no-thinking ability. In the second stage, we collect rollouts with the hybrid reasoning strategy (determining to use the thinking or no-thinking prompts according to the action entropy and threshold) and solely train thinking ability using reasoning data in the rollouts. In case of forgetting abilities trained in the Stage I, we add a KL regularization to maintain a proper distance between the updated model and the stage I's checkpoint. The whole training objective is summarized as follows: 

\begin{align}
L(\theta)&= \mathbb{E}_l [ \min ( \rho_{l}(\theta) \hat{A}_l, \notag \\&\mathrm{clip}\left(\rho_{l}(\theta)\hat{A}_l, 1-\epsilon, 1+\epsilon\right) \hat{A}_l) ] \times \mathbb{I}_{Tk}(\bm{y})) \notag \\&- \beta \mathbb{D}_{KL}\left[\pi_{\theta}(\bm{y}|\bm x) || \pi_{\theta_{\mathrm{ref}}}(\bm{y}|\bm x)\right] \times \left(1-\mathbb{I}_{Tk}(\bm{y})\right)
\label{eq:ppo_clip_objective}
\end{align}
where $\rho_{l}(\theta) = \frac{\pi_{\theta}(y_l|\bm{y}_{<l},\bm x)}{\pi_{\theta_{\mathrm{old}}}(y_l|\bm{y}_{<l},\bm x)}$,
$\mathbb{I}_{Tk}(\bm{y}) = 1$ if $\bm{y}$ is generated in thinking mode else 0, $\hat{A}_l$ is the Generalized Advantage Estimation (GAE). In stage I, $\beta$ is set to 0 and $\mathbb{I}_{Tk}(\bm{y})=0$ since we collect rollouts using the no-thinking prompt. In Stage II, we set $\beta$ to 0.1.

\begin{table*}[t]
\centering
\caption{\textbf{Comparison with state of the art.} SR = success rate, SEL = success rate weighted by episode length, Ours = our proposed hybrid reasoning strategy, NRD = no-reasoning dataset, HRD = hybrid-reasoning dataset, TF = training-free. $K$ in the Thinking-Every-$K$-Steps baseline and the size of no-thinking window (NTW) are equally set to 5 for a fair comparison. $\checkmark$/$\times$ refers to the corresponding annotated semantic map (ASM) included/not included in the input of agents. \#Token/E and \#Token/S denote the number of tokens generated by the agent per episode and per step, respectively. RL$_{\text{I}}$ and RL$_{\text{II}}$ refer to the two stages of online RL training.}
\label{tab:main}
\resizebox{\linewidth}{!}{%
\begin{tabular}{l|l|lcc|cccc} 
\toprule
\textbf{Training Recipe}                                                                                                                                                   & \textbf{Model Name}                                                                                                      & \textbf{Reasoning Strategy}                                                                                   & \textbf{ASM}                                                                                                      & \textbf{Dataset}                                                                                           & \textbf{SR\%}$\uparrow$                                                                                               & \textbf{SEL\%}$\uparrow$                                                                                             & \textbf{\#Token/E}$\downarrow$                                                                                               & \textbf{\#Token/S}$\downarrow$                                                                                      \\ 
\midrule
\multirow{3}{*}{TF}                                                                                                                                                        & GPT-4o \cite{gpt4o}                                                                                                             & No-Thinking                                                                                                   & $\checkmark$                                                                                                        & -                                                                                                          & 8.0                                                                                                         & 4.4                                                                                                         & $\scinot{5.6}{2}$                                                                                                  & 3.1                                                                                                         \\
                                                                                                                                                                           & o3 \cite{o3}                                                                                                                   & Dense-Thinking                                                                                                & $\checkmark$                                                                                                        & -                                                                                                          & 27.0                                                                                                        & 12.3                                                                                                        & $\scinot{2.4}{5}$                                                                                                  & 539.4                                                                                                       \\
                                                                                                                                                                           & Gemini2.5-Pro \cite{gemini2.5pro}                                                                                                       & Dense-Thinking                                                                                                & $\checkmark$                                                                                                        & -                                                                                                          & 27.5                                                                                                        & 16.2                                                                                                        & $\scinot{2.3}{5}$                                                                                                  & 804.8                                                                                                       \\ 
\midrule
\multirow{3}{*}{SFT}                                                                                                                                                       & NaVILA \cite{navila}                                                                                                              & No-Thinking                                                                                                   & $\times$                                                                                                            & NRD                                                                                                        & 30.5                                                                                                        & 12.1                                                                                                        & $\scinot{1.8}{3}$                                                                                                  & 3.7                                                                                                         \\ 
\cmidrule{2-9}
                                                                                                                                                                           & \multirow{2}{*}{Qwen2.5VL-3B \cite{qwen25vl}}                                                                                            & No-Thinking                                                                                                   & $\times$                                                                                                            & NRD                                                                                                        & 36.5                                                                                                        & 28.8                                                                                                        & $\scinot{1.5}{3}$                                                                                                  & 3.8                                                                                                         \\
                                                                                                                                                                           &                                                                                                                          & No-Thinking                                                                                                   & $\checkmark$                                                                                                        & NRD                                                                                                        & 50.0                                                                                                        & 41.2                                                                                                        & $\scinot{1.2}{3}$                                                                                                  & 3.8                                                                                                         \\ 
\cmidrule{1-9}
\multirow{4}{*}{H-SFT}                                                                                                                                                    & \multirow{4}{*}{Qwen2.5VL-3B \cite{qwen25vl}}                                                                                            & No-Thinking                                                                                                   & $\checkmark$                                                                                                        & HRD                                                                                                        & 49.5                                                                                                        & 39.1                                                                                                        & $\scinot{1.2}{3}$                                                                                                  & 3.8                                                                                                         \\
                                                                                                                                                                           &                                                                                                                          & Thinking-Every-$K$-Steps                                                                                      & $\checkmark$                                                                                                        & HRD                                                                                                        & 60.0                                                                                                        & 40.3                                                                                                        & $\scinot{4.8}{3}$                                                                                                  & 17.6                                                                                                        \\
                                                                                                                                                                           &                                                                                                                          & Dense-Thinking                                                                                                & $\checkmark$                                                                                                        & HRD                                                                                                        & 21.0                                                                                                        & 8.2                                                                                                         & $\scinot{3.6}{4}$                                                                                                  & 72.1                                                                                                        \\
                                                                                                                                                                           &                                                                                                                          & {\cellcolor[rgb]{0.929,0.949,0.988}}\cellcolor[rgb]{0.929,0.949,0.988}\cellcolor[HTML]{EDF2FC}\textbf{Ours} & {\cellcolor[rgb]{0.929,0.949,0.988}}\cellcolor[rgb]{0.929,0.949,0.988}\cellcolor[HTML]{EDF2FC}\textbf{$\checkmark$} & {\cellcolor[rgb]{0.929,0.949,0.988}}\cellcolor[rgb]{0.929,0.949,0.988}\cellcolor[HTML]{EDF2FC}\textbf{HRD} & {\cellcolor[rgb]{0.929,0.949,0.988}}\cellcolor[rgb]{0.929,0.949,0.988}\cellcolor[HTML]{EDF2FC}\textbf{59.5} & {\cellcolor[rgb]{0.929,0.949,0.988}}\cellcolor[rgb]{0.929,0.949,0.988}\cellcolor[HTML]{EDF2FC}\textbf{48.9} & {\cellcolor[rgb]{0.929,0.949,0.988}}\cellcolor[rgb]{0.929,0.949,0.988}\cellcolor[HTML]{EDF2FC}$\bm{2.3 \times 10^{3}}$ & {\cellcolor[rgb]{0.929,0.949,0.988}}\cellcolor[rgb]{0.929,0.949,0.988}\cellcolor[HTML]{EDF2FC}\textbf{9.2}  \\ 
\midrule
\multirow{2}{*}{SFT+RL$_\text{I}$}                                                                                                                                         & NaVILA \cite{navila}                                                                                                               & No-Thinking                                                                                                   & $\times$                                                                                                            & NRD                                                                                                        & 44.0                                                                                                        & 27.1                                                                                                        & $\scinot{1.5}{3}$                                                                                                  & 3.8                                                                                                         \\
                                                                                                                                                                           & Qwen2.5VL-3B \cite{qwen25vl}                                                                                                      & No-Thinking                                                                                                   & $\checkmark$                                                                                                        & NRD                                                                                                        & 70.5                                                                                                        & 52.5                                                                                                        & $\scinot{8.1}{2}$                                                                                                  & 3.7                                                                                                         \\ 
\midrule
\rowcolor[rgb]{0.929,0.949,0.988} {\cellcolor[rgb]{0.929,0.949,0.988}}                                                                                                     & {\cellcolor[rgb]{0.929,0.949,0.988}}                                                                                     & \cellcolor[HTML]{EDF2FC}No-Thinking                                                                           & $\checkmark$                                                                                                        & HRD                                                                                                        & 70.0                                                                                                        & 48.8                                                                                                        & $\scinot{8.2}{2}$                                                                                                  & 3.6                                                                                                         \\
\rowcolor[rgb]{0.929,0.949,0.988} {\cellcolor[rgb]{0.929,0.949,0.988}}                                                                                                     & {\cellcolor[rgb]{0.929,0.949,0.988}}                                                                                     & \cellcolor[HTML]{EDF2FC}Thinking-Every-$K$-Steps                                                              & $\checkmark$                                                                                                        & HRD                                                                                                        & 78.5                                                                                                        & 46.7                                                                                                        & $\scinot{3.5}{3}$                                                                                                  & 17.3                                                                                                        \\
\rowcolor[rgb]{0.929,0.949,0.988} {\cellcolor[rgb]{0.929,0.949,0.988}}                                                                                                     & {\cellcolor[rgb]{0.929,0.949,0.988}}                                                                                     & \cellcolor[HTML]{EDF2FC}Dense-Thinking                                                                        & $\checkmark$                                                                                                        & HRD                                                                                                        & 34.5                                                                                                        & 15.6                                                                                                        & $\scinot{3.1}{4}$                                                                                                  & 70.4                                                                                                        \\
\rowcolor[rgb]{0.929,0.949,0.988} \multirow{-4}{*}{\cellcolor[HTML]{EDF2FC} H-SFT+RL$_\text{I\&II}$} & \multirow{-4}{*}{{\cellcolor[rgb]{0.929,0.949,0.988}}\cellcolor[rgb]{0.929,0.949,0.988}\cellcolor[HTML]{EDF2FC}HiRO-Nav} & \textbf{Ours}                                                                                               & \textbf{$\checkmark$}                                                                                               & \textbf{HRD}                                                                                               & \textbf{81.0}                                                                                               & \textbf{57.2}                                                                                               & $\bm{2.7 \times 10^{3}}$                                                                                               & \textbf{13.5}                                                                                               \\
\bottomrule
\end{tabular}%
}
\end{table*}
\section{Experimental Results}
\subsection{Implementation Details}

\textbf{H-SFT}. The short-term memory size $w$ is set to 4. To implement the ASM, we use ground truth object locations and the depth sensor from the AI2-Thor simulator \cite{ai2thor}. These can be replaced by advanced deep models such as Mask R-CNN \cite{he2017mask} and Depth-Anything \cite{depthanything}. We finetune only the LLM parameters on the HRD for 1 epoch. The training batch size is set to 256, and the learning rate is set to 2e-5. 
\\
\textbf{Online RL}. Following \citet{poliformer}, we use the ProcThor-150k houses with ~40k annotated Objaverse 3D assets and the same reward setting. The actor model is initiated from the hybrid fine-tuned VLM. To implement the value network, we initiate from the same VLM and apply a linear layer taking the hidden state in the VLM's last layer as input to predict values. During training, the rollout size is set to 48, and the PPO update mini-batch size is set to 384. The maximum environment interaction is set to 300 to increase rollout collection efficiency. In each training stage, we train the model for 10 steps and select the checkpoint with the highest rollout success rate. We use Verl-Agent \citep{gigpo} as our training framework. 
\subsection{Evaluation Setup}

\textbf{Evaluation}. We perform evaluation on the \textsc{CHORES}-$\mathbb{S}$ ObjectNav benchmark\citep{spoc} which contains 200 tasks in 200 scenes, with a Stretch RE-1 robot \cite{poliformer} setting. Following \citep{poliformer}, the maximum interaction during evaluation is 600. We choose \textit{Success Rate} (SR) and \textit{Success weighted by Episode Length} (SEL) as the metrics to evaluate performance. The SEL score is defined as follows:
\begin{equation}
    SEL = \frac{1}{N}\sum_{i=1}^{N}S_{i}\frac{w_i}{max(w_i, e_i)}
\end{equation}
where $S_i$ is a binary indicator of success for episode $i$, $w_i$ is the shortest number of steps to find the target, $e_i$ is the number of steps taken by the agent.  The entropy threshold $\tau$ is set to 0.6 and the no-thinking window size is set to 5 across our main experiments.
\\
\textbf{Baselines}. We evaluate HiRO-Nav against three categories of baseline, each distinguished by distinct reasoning strategies: (1) \textbf{No-Thinking}, (2) \textbf{Thinking-Every-$K$-Steps}, as introduced in OctoNav \citep{octonav}, and (3) \textbf{Dense-Thinking}, which performs thinking at every step. For each category, we include a variant of our trained model employing the respective reasoning strategies. Additionally, we compare our approach with powerful general VLMs, namely GPT-4o \citep{gpt4o}, o3 \citep{o3}, and Gemini2.5-Pro \citep{gemini2.5pro}. Due to the intrinsic internal reasoning mechanism of o3 and Gemini2.5-Pro, we take them as dense-thinking baselines. In contrast, GPT-4o is taken as a no-reasoning baseline. 

\begin{figure}[t]
    \centering
    \centering
    \includegraphics[width=0.8\linewidth]{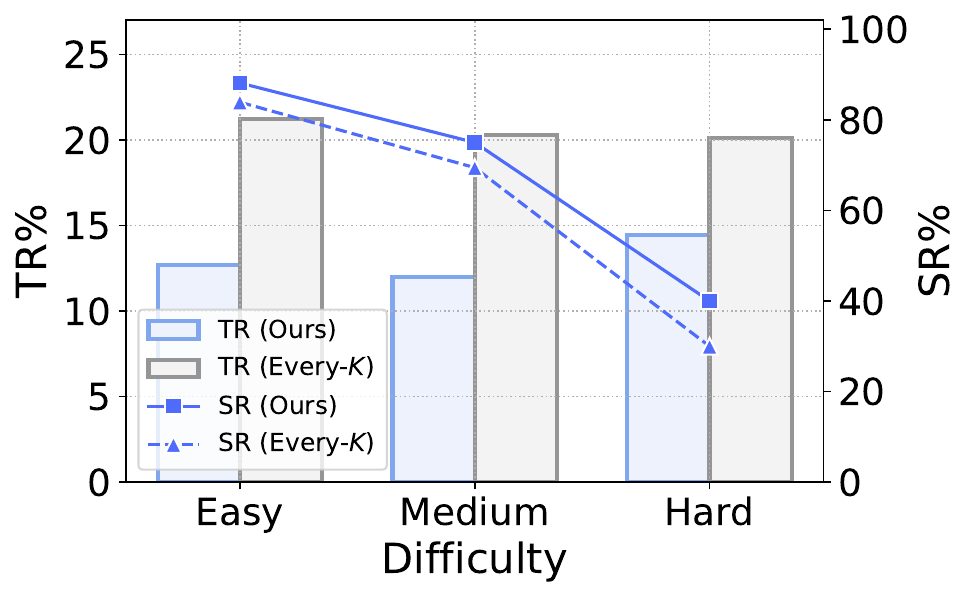}
    \caption{\textbf{Comparison of reasoning efficiency of hybrid reasoning (Ours) and thinking-every-$\bm K$-steps (Every-$\bm K$)}. We divide the navigation tasks into different difficulty levels based on the ground truth shortest path lengths. Our hybrid reasoning method consistently outperforms the baseline reasoning approach across all difficulty levels while maintaining a lower thinking ratio. TR=Thinking Ratio. SR= Success Rate.}
    \label{fig:step_meta}
\end{figure}
\begin{figure*}[ht]
  \centering
  \begin{subfigure}[t]{0.31\linewidth}
    \centering
    \includegraphics[width=\linewidth]{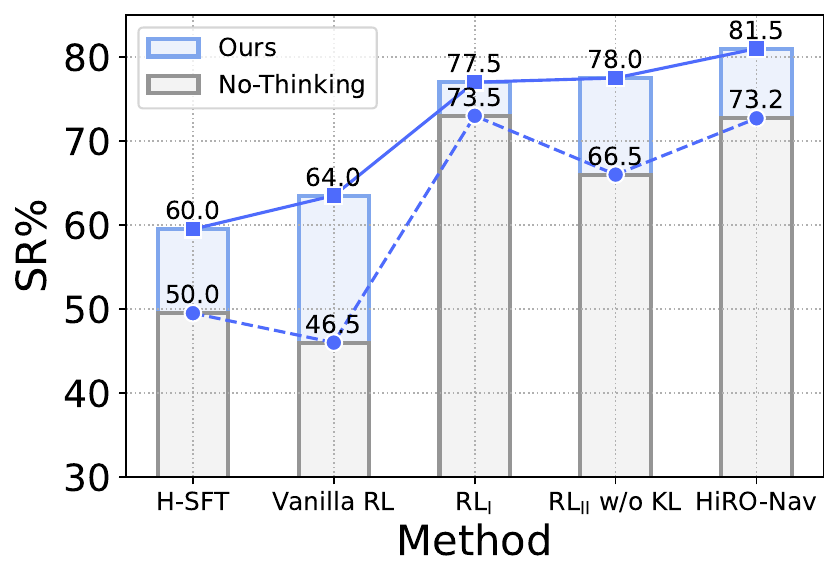}
    \caption{\textbf{Ablation study on two-stage online RL.} Vanilla RL refers to directly optimizing hybrid reasoning ability during RL training.}
    \label{fig:ablation}
  \end{subfigure}
  \quad
  \begin{subfigure}[t]{0.31\linewidth}
    \centering
    \includegraphics[width=\linewidth]{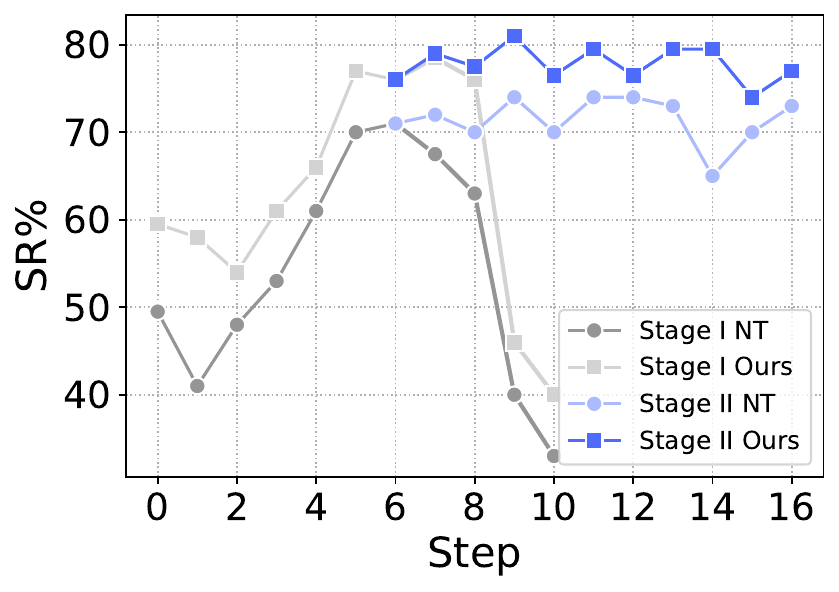}
    \caption{\textbf{Navigation ability dynamics in no-thinking and hybrid-thinking mode(Ours) during online RL}. NT=No-Thinking. \looseness-1}
    \label{fig:rl_curve}
  \end{subfigure}
  \quad
  \begin{subfigure}[t]{0.31\linewidth}
    \centering
     \includegraphics[width=\linewidth]{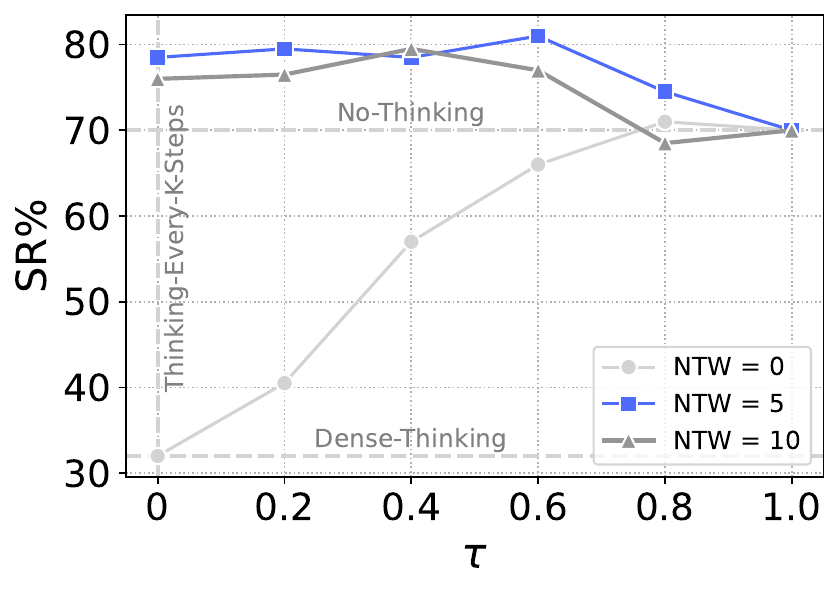}
    \caption{\textbf{Hyperparameter analysis on AE threshold $\tau$ and no-thinking window (NTW) size.} Our method degrades to thinking-evey-$K$-steps with $K$=NTW when $\tau$=0. \looseness-1}
    \label{fig:hypercurve}
  \end{subfigure}
  \caption{\textbf{Ablation study.} \textbf{Fig. (a)\&(b)}: Vanilla RL fails to effectively enhance hybrid reasoning ability due to a decline in no-thinking ability. In contrast, our two-stage training paradigm successfully improves the agent's no-thinking ability in Stage I and maintained by KL constraint in Stage II, which subsequently enhance the hybrid reasoning ability. \textbf{Fig. (c)}: The superior performance of NTW\textgreater0 highlights its effectiveness. Our hybrid reasoning enables the agent to achieve improved performance with a reduced thinking ratio (indicated by high $\tau$), reaching a SOTA result at $\tau$=0.6 and NTW=5.}
\end{figure*}
\subsection{Main Results}
\textbf{Overall Performance.} As shown in Table \ref{tab:main}, HiRO-Nav with hybrid-reasoning achieves a sota trade-off between navigation success rate and token efficiency, outperforming all other baselines in terms of success rate and SEL, while maintaining an efficient token cost that is significantly lower than dense-thinking and comparable to no-thinking. 
\\
\textbf{Effect of Reasoning Strategies.} For the variants of our HiRO-Nav, no-thinking outperforms dense-thinking by 35\%, verifying that overthinking adversely affects model performance. Thinking-every-$K$-steps improves upon no-thinking by 7.5\%, mitigating overthinking by reducing thinking frequency. Our Hybrid reasoning strategy further surpasses thinking-every-$K$-steps by an additional 4.5\%, verifying the effectiveness of our design. 
Furthermore, we compare our hybrid reasoning strategy with thinking-every-$K$-steps through a fine-grained analysis stratified by task difficulty on success rate. We measure the task difficulty by the shortest path length. We split tasks into `Easy' (\textless 50 steps), `Medium' (50$\sim$150 steps) and `Hard'(\textgreater150 steps). As shown in Fig.~\ref{fig:step_meta}, our hybrid reasoning method consistently outperforms the thinking-every-$K$-steps method across all difficulty levels while maintaining a 10\% lower thinking ratio. This demonstrates that our reasoning strategy more effectively incentivizes the LRM’s reasoning ability by invoking reasoning at more appropriate times, achieving higher performance and lower reasoning cost.\\
\textbf{Effect of the Training Pipeline.} We analyze the effect of our training pipeline, which includes HRD construction, hybrid supervised fine-tuning, and online reinforcement learning. For the HRD, we demonstrate that incorporating ASM as memory improves the agent’s navigation ability. Specifically, Qwen2.5VL-3B fine-tuned with ASM on the same NRD outperforms the model without ASM by 13.5\% in SR. The HRD further equips the model with thinking capability without compromising no-thinking performance, as evidenced by the improvements of the hybrid fine-tuned Qwen2.5VL-3B on HRD with both hybrid reasoning and thinking-every-$K$-steps strategies compared to fine-tuned on NRD alone. Through online RL, HiRO-Nav surpasses the fine-tuned model by 10\% to 20\% across all reasoning strategies, demonstrating its effectiveness. \looseness-1

\subsection{Ablation Study}



\textbf{Effect of the two-stage online RL.} In Fig.~\ref{fig:ablation}, we observe the following: (1) Directly optimizing hybrid reasoning ability is suboptimal. The no-thinking ability collapses, exhibiting a 3.5\% performance drop compared to the hybrid supervised fine-tuning (H-SFT) model, which ultimately limits the overall hybrid reasoning capability. In contrast, training the two modes separately shows promising results. In Stage I, we only train the no-thinking ability, which improves performance by 19.5\%. Simultaneously, the hybrid reasoning performance also increases by 17.5\%. After training the thinking ability in Stage II, the hybrid reasoning achieves an additional 4\% performance gain, verifying the effectiveness of our two-stage RL training paradigm. (2) In Stage II, we further investigate the impact of the KL penalty. Disabling the KL penalty leads to a 3\% performance drop in the no-thinking ability compared to Stage I, and a 4\% drop compared to using the activated KL penalty, highlighting the importance of the KL penalty in preventing mode collapse. \looseness-1

We further visualize the dynamics of navigation ability during the two-stage reinforcement learning process in Fig.~\ref{fig:rl_curve}.
HiRO-Nav with hybrid reasoning consistently outperforms that with no-thinking, validating the efficacy of our hybrid reasoning design.
In Stage I, the navigation ability initially improves rapidly but experiences a decline after step 6. We attribute this drop to the model converging to a conservative policy, such as frequently moving backward to avoid collisions. In Stage II, we successfully prevent the collapse observed in the no-thinking mode, which in turn facilitates the improvement of the thinking mode. \\
\textbf{Effect of action entropy threshold $\bm\tau$ and no-thinking window size (NTW)}. We perform analysis using HiRO-Nav with the hybrid reasoning strategy. As illustrated in Fig.~\ref{fig:hypercurve}, we observe that, (1) under $\text{NTW=0}$ setting, the agent struggles to outperform the no-thinking baseline, indicating that the entropy trap significantly affects our hybrid reasoning strategy. In contrast, when $\text{NTW}>0$, the agent significantly outperforms the no-thinking baseline, verifying the effectiveness of no-thinking window. (2) The agent's performance initially improves  as $\tau$ increases, demonstrating that our method achieves better results with a reduced thinking ratio. This highlights the importance of thinking at appropriate times to fully unleash the agent’s reasoning ability. When $\tau$ approaches 1.0,  the agent stops thinking, and the performance eventually converges to the no-thinking baseline. 
\begin{table}[t]
\centering
\caption{\textbf{Robustness analysis of HiRO-Nav using annotated semantic maps constructed by deep models (DM) in place of ground truth (GT) ASMs.} We apply Depth-Anything\cite{depthanything} to estimate depth information and a fine-tuned Mask R-CNN \cite{he2017mask} to perform instance segmentation during ASM construction. Despite of performance drop due to noise introduced by inaccurate estimation, HiRO-Nav with hybrid reasoning still outperforms no-thinking with ground truth ASMs, showing its robustness in resisting noise in ASM.}
\label{tab:my-table}
\resizebox{0.7\linewidth}{!}{%
\begin{tabular}{@{}c|l|cc@{}}
\toprule
\textbf{ASM}        & \textbf{Reasoning Strategy} & \textbf{SR\%} & \textbf{SEL\%} \\ \midrule
\multirow{3}{*}{GT} & Dense-Thinking              & 34.5          & 15.6           \\
                    & No-Thinking                 & 70.0          & 48.8           \\
                    & Ours                        & 81.0          & 57.2           \\ \midrule
DM                  & Ours                        & 72.0          & 46.8           \\ \bottomrule
\end{tabular}%
}
\label{tab:robustness}
\end{table}

\subsection{Robustness Analysis.}
We further verify the robustness of HiRO-Nav by replacing the ground truth ASMs with ASMs constructed using deep learning models. Specifically, we apply the Depth-Anything \cite{depthanything} model to estimate depth information and use a finetuned Mask R-CNN \cite{he2017mask} model for instance segmentation\footnote{We follow the Mask R-CNN fine-tuning scripts in ALFWord \cite{alfworld} official repo.}. As shown in Table \ref{tab:robustness}, our agent experiences only a slight performance drop due to the reduced quality of the estimated ASMs, while still maintaining strong navigation capabilities. Moreover, our agent continues to outperform the no-thinking and dense-thinking baselines, further demonstrating that even with lower-quality ASMs, the hybrid reasoning strategy still can effectively incentivize the LRM's reasoning capabilities, verifying the robustness of our HiRO-Nav. \looseness-1

\subsection{Navigation Ability Upper Bound.} 
\begin{figure}[t]
     \centering
    \includegraphics[width=0.8\linewidth]{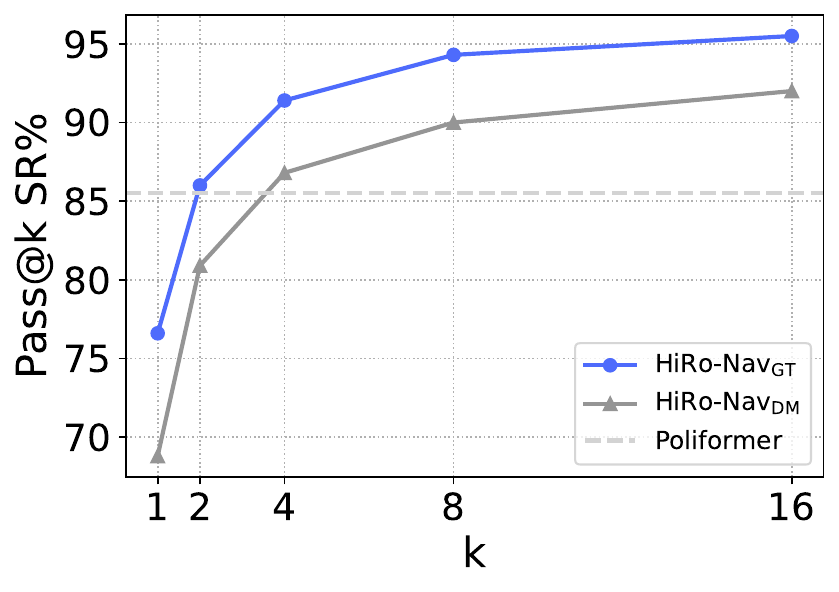}
    \caption{\textbf{Pass@k curves of HiRO-Nav with hybrid reasoning.} GT and DM refer to the ground truth ASMs and the ASMs estimated by deep models as in Tab.~\ref{tab:robustness} .We evaluation for 16 times with temperature=0.2. The navigation ability upper bound of HiRO-Nav outperforms task-specific sota method Poliformer\cite{poliformer}, even when using noisy deep model estimated ASMs.}
    \label{fig:passk}
\end{figure}
We further explore the HiRO-Nav performance upper bound by performing Pass@k evaluation \cite{passk} with the hybrid reasoning strategy. We evaluate using both ground truth ASMs (HiRO-Nav$_{\text{GT}}$) and ASMs estimated by deep models (HiRO-Nav$_{\text{DM}}$) for 16 times each with temperature=0.2 and calculate Pass@1,2,4,8,16 success rate. As shown in Fig.~\ref{fig:passk}, with either ASM,  the navigation success rate increases as k increases, reaching a plateau of 95.5\% for HiRO-Nav$_{\text{GT}}$ and 92.0\% for HiRO-Nav$_{\text{DM}}$ when k=16. HiRO-Nav's performance upper bound significantly exceeds that of the task-specific state-of-the-art RL method Poliformer \cite{poliformer}, demonstrating the strong navigation capabilities of HiRO-Nav. \looseness-1

\section{Conclusion}
To mitigate overthinking and intelligently incentivize LRM's reasoning capability in navigation,  we develop the first kind of navigation agent HiRO-Nav with a hybrid reasoning capability. By analyzing action entropy patterns,  we propose a hybrid reasoning strategy where the agent only performs thinking for high-entropy actions. Through a tailored training pipeline containing a hybrid supervised fine-tuning and a two-stage online RL with the hybrid reasoning strategy,  our HiRO-Nav achieves a SOTA trade-off between navigation performance and reasoning efficiency compared with existing reasoning strategies. 

Despite these advancements, our method currently relies on a predefined entropy threshold and a reactive process of regenerating CoT traces after an initial action is predicted. These factors introduce latency and limit overall efficiency. In future work, we aim to train the agent to autonomously decide when to reason based on action entropy patterns, creating a more adaptive and efficient system.
{
    \small
    \bibliographystyle{ieeenat_fullname}
    \bibliography{main}
}
\newpage
\appendix
\onecolumn
\clearpage
\setcounter{page}{1}

\section{Related Work}
\label{sec:app_related_work}
\subsection{Foundation Models as Navigation Agents}
LRMs have been introduced to handle navigation tasks~\citep{navila,uni-navid,octonav,sgnav,mem2ego,topv-nav} due to their rich prior knowledge and the ability to resolve problems in complex environments.  Zero-shot navigation agents like VLFM \citep{vlfm}, SG-Nav \citep{sgnav}, and VLMNav~\citep{vlmnav} leverage complex input prompts to incentivize the planning ability of LLMs or LVLMs. The upper bound of these methods' capabilities is constrained by their training-free approach. Early finetuned large navigation agents are trained based on the historical observations extracted from expert trajectories \citep{navid,uni-navid,navila}. MapNav \citep{mapnav} collects annotated semantic map data as an additional input modality to train a multi-modal model. Recent works further improve navigation ability by training models on a collected reasoning dataset \citep{aux-think,vln-r1,divscene}.For example,  OctoNav \citep{octonav} collects a Think-Before-Action dataset based on expert trajectories on multiple navigation tasks, and further trains the model through supervised finetuning and reinforcement finetuning methods \citep{grpo, ppo}. 

\subsection{Hybrid or Adaptive Reasoning of LRMs}
LRMs like Deepseek-R1 \cite{deepseekr1} have made promising achievements in complex reasoning tasks by using CoT \cite{cot}. However, existing studies have pointed out that overthinking on simple tasks may hinder performance and is inefficient \cite{think-or-not-think,cot-or-not, overthinking,yue2025don}. Recent works incentivize models to adaptively adjust their reasoning length or switch between different reasoning modes guided by carefully designed reinforcement fine-tuning\cite{when-to-think,adactrl,adaptthink,lhrms,wang2025adaptive}. 
However, when to think in navigation tasks still remains underexplored. OctoNav \citep{octonav} performs CoT reasoning every k steps during testing. Aux-Think \cite{aux-think} proposes to sft the model with a mixture of reasoning data and action-only data, while outputting actions only when testing. However, these existing methods engage in reasoning by predefined rules regardless of navigation dynamics, which will hinder model performance eventually. To this end, we are the first to propose a navigation agent with hybrid reasoning ability adaptively determining whether to perform thinking to achieve better navigation performance and efficiency.

\section{Additional Details of ObjectNav Task Settings}\label{sec:action_space}
\begin{figure}[h]
    \centering
    \includegraphics[width=0.5\linewidth]{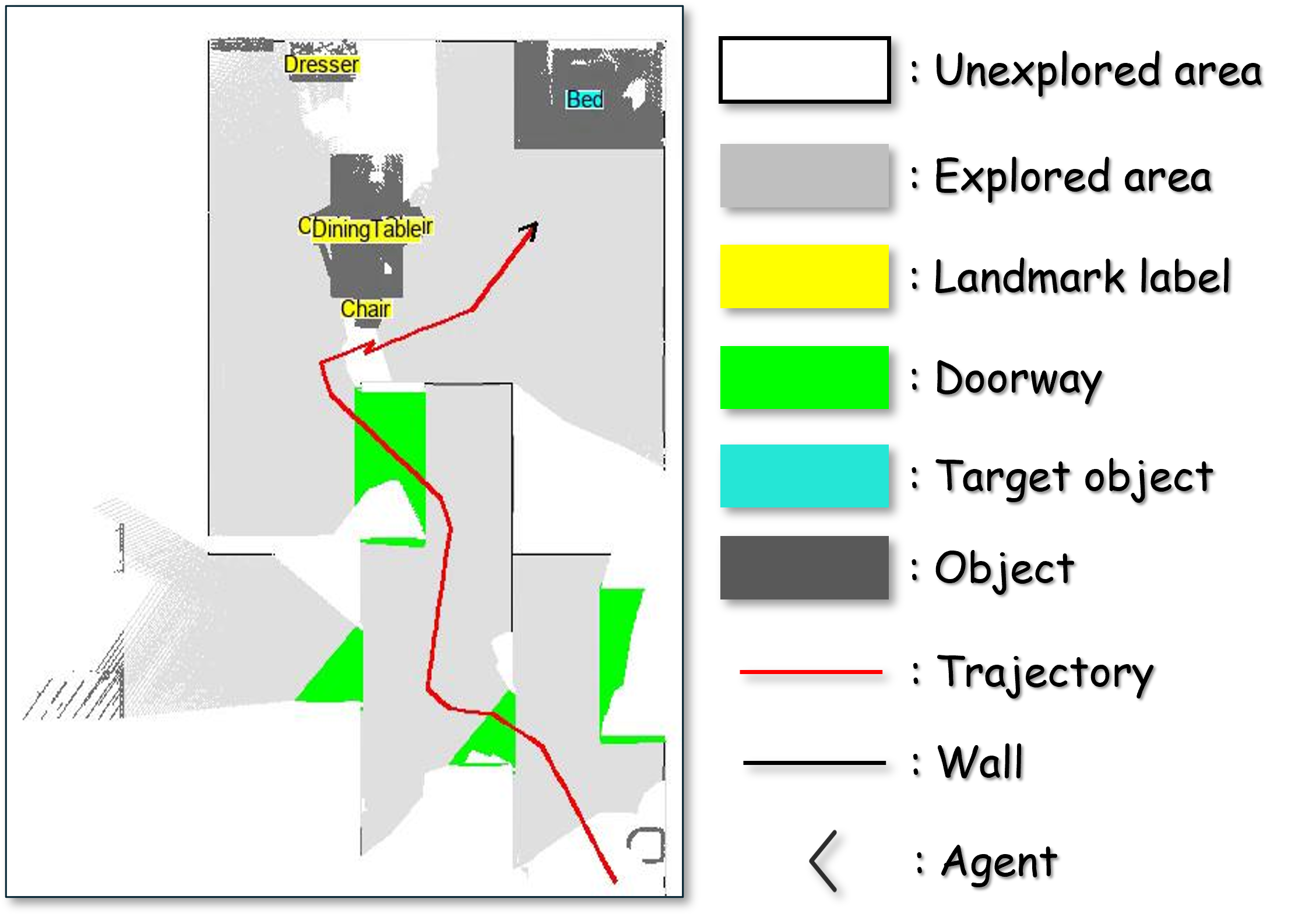}
    \caption{An example of annotated semantic map.}
    \label{fig:map}
\end{figure}
\noindent \textbf{Task Success Condition.}The task is considered successful if the agent terminates navigation by emitting the ``end'' action within a specified step limit, and the target object is within the agent’s view and within a certain distance from its current location.
\\
\textbf{Action Space}. We provide details of the action space the RE-Strech 1 robot in Table \ref{tab:actions}. 
\begin{table}[h] 
\centering
\caption{Action Space and corresponding arguments. All actions are in textual form.}
\label{tab:actions}
\begin{tabular}{cc}
\toprule
\textbf{Action}               & \textbf{Argument}                \\ \midrule
move\_ahead          & 0.2 meter \\
move\_back           & 0.2 meter    \\
rotate\_left         & 30 degree  \\
rotate\_right        & 30 degree \\
rotate\_left\_small  & 5 degree   \\
rotate\_right\_small & 5 degree  \\
end                  &  \textbackslash{}      \\ \bottomrule
\end{tabular}

\end{table}
\\
\textbf{Reward Shaping}. We use the same reward setting as in Poliformer \citep{poliformer} during the reinforcement learning stage. Specifically, the total reward is defined as $\mathcal{R} = \mathcal{R}_{penalty}+\mathcal{R}_{success}+\mathcal{R}_{distance}$, where $\mathcal{R}_{penalty}$ is a step penalty and set to -0.01, to encourage efficient navigation, $\mathcal{R}_{success}$ is set to 10 when the agent successfully completes the task and 0 otherwise, and $\mathcal{R}_{distance}$ represents the change in L2 distance at the current step; it is equal to the positive distance reduction if the agent moves closer to the target and 0 otherwise, instead of a negative value, to encourage exploration \looseness-1.

\section{Additional Details in Methodology}\label{sec:app_method}
\begin{figure*}[htbp]
    \centering
    \includegraphics[width=\textwidth]{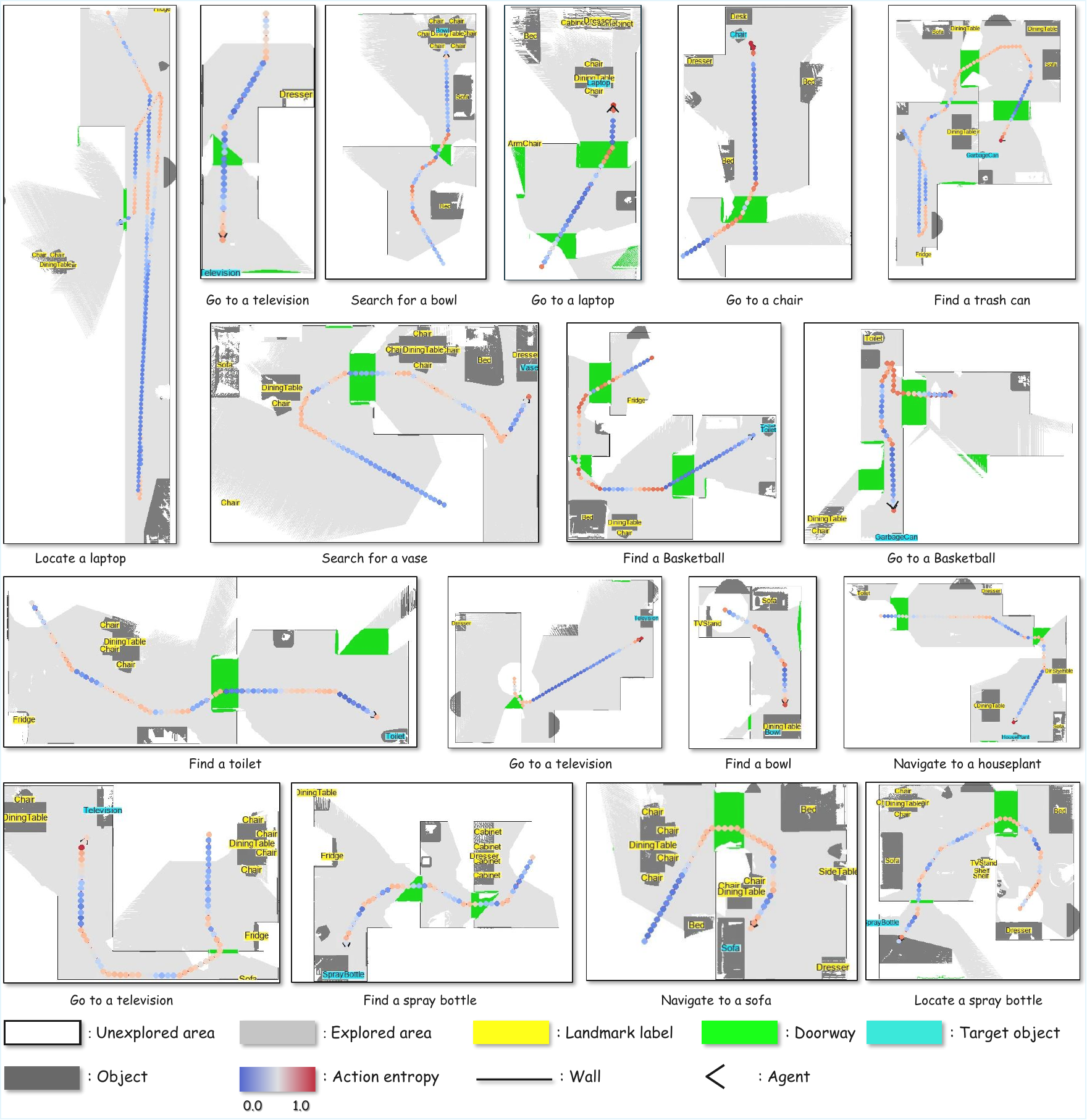}
    \caption{Additional visualization examples of action entropy at each navigation waypoint.}
    \label{fig:entropy_example_app}
\end{figure*}

\begin{figure*}[h]
    \centering
    \begin{subfigure}[b]{0.48\textwidth}
    \centering
    \includegraphics[width=\textwidth]{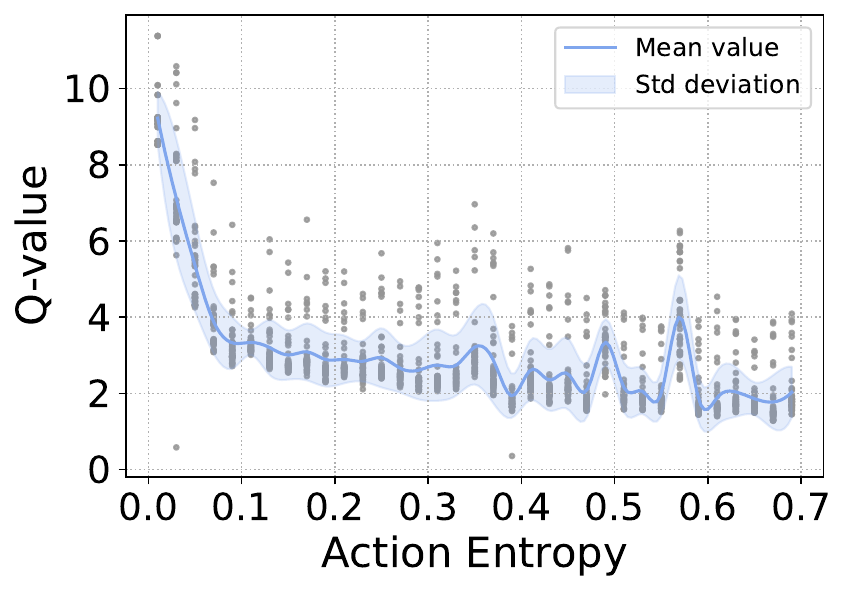}
     \caption{First token entropy as action entropy. }
  \end{subfigure}
  \quad
  \begin{subfigure}[b]{0.48\textwidth}
    \centering
    \includegraphics[width=\textwidth]{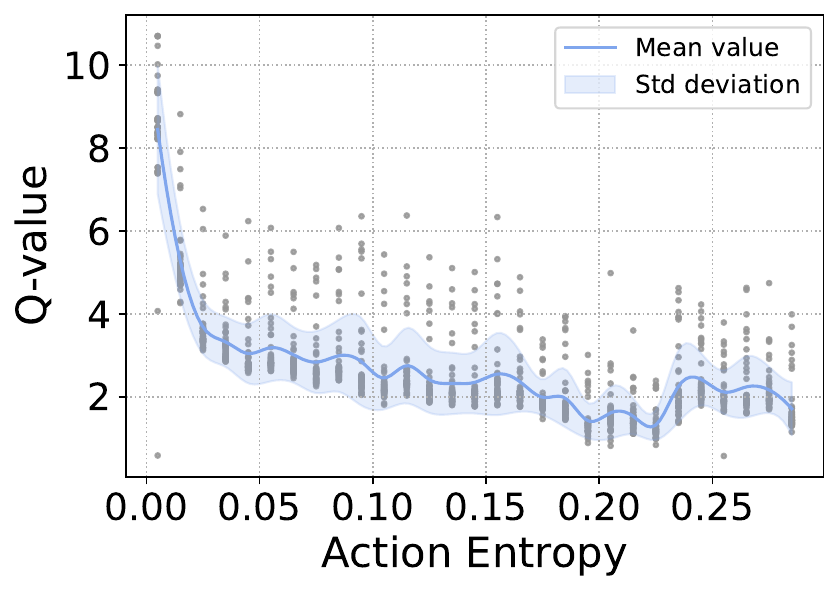}
     \caption{Average token entropy as action entropy. }
  \end{subfigure}
\caption{Two action entropy calculation method show the same relationship between action entropy and Q-value. }
\label{fig:two_method}
\end{figure*}

\subsection{Action Entropy} \label{sec:action_entropy}
In Section \ref{sec:critical_step}, we analyze how action entropy evolves over navigation processes. To calculate the action entropy, we compare two different methods: first-token entropy and mean token entropy as action entropy. We evaluate using a VLM fine-tuned on expert trajectories on the top 50 difficult tasks on the benchmark. As shown in Fig. \ref{fig:two_method} and Table \ref{tab:mean_first_ent}, both methods capture the same relationship with the Q-value, and their performances are similar. Given that each action consists of a maximum of 4 tokens and considering the auto-regressive generation pattern of LLMs, we argue that using the first-token entropy to represent action entropy is an efficient choice. Based on the action entropy analysis, we identify that high entropy actions posing a significant impact on navigation. Here we provide additional examples in Fig. \ref{fig:entropy_example_app} to support this claim.

\begin{table}[h]
\centering
\caption{We compare using first-token entropy versus mean token entropy as measures of action entropy. The performance difference between the two methods is marginal.}
\label{tab:mean_first_ent}
\begin{tabular}{@{}c|c|cc@{}}
\toprule
\textbf{Model}            & \textbf{Action Entropy} & \textbf{SR\%} & \textbf{SEL\%} \\ \midrule
\multirow{2}{*}{Qwen2.5VL-3B$_{\text{H-SFT}}$}      & First Token             & 59.5                     & 48.9            \\
                          & Mean              & 58.0                     & 43.1            \\ \midrule
\multirow{2}{*}{HiRO-Nav} & First Token             & 81.0                     & 57.2            \\
                          & Mean              & 79.5                     & 53.1            \\ \bottomrule
\end{tabular}
\end{table}


\subsection{Annotated Semmantic Map Construction}
\label{sec:semantic map}

To maximize the visual information within the limited context length of VLMs, we construct annotated semantic maps \citep{mapnav} to serve as the agent’s memory. Specifically, our annotated semantic map encodes explored areas, the agent’s trajectory, and object locations with corresponding annotations. An example is given in the Fig. \ref{fig:map}.
To reduce redundancy caused by overlapping objects, only furniture items and the target items are annotated as landmark objects on the map (e.g., dining table, bed). These landmark objects provide spatial references that assist the VLM during planning. Object segmentation and recognition are performed by an external module; in this work, we utilize ground-truth feedback from the AI2-THOR simulator \citep{ai2thor}. In real-world applications, this can be replaced by advanced segmentation and detection methods, such as Mask R-CNN \citep{he2017mask}.
\subsection{Reasoning Traces Collection}
\label{sec:hrd_app}
We first filter data with top 20\% high action entropy and apply Gemini-2.0-flash \cite{gemini} to generate reasoning traces. 
Prior work \citep{ui-tars} has highlighted that when the ground-truth answer is available, language models may neglect the internal logical process of the problem, resulting in reasoning traces that overfit to the answer. 
To mitigate this issue, we employ the same thought bootstrapping strategy by using Gemini-2.0-flash to iteratively generate reasoning traces until one is found whose final answer matches the ground truth action. If the number of attempts exceeds the allowed maximum, we mark these data as counterintuitive and do not generate reasoning traces for them.
To further enhance the quality of the reasoning data, we further filter out data of which answers are inconsistent with the reasoning traces, and the reasoning traces containing incorrect visual information to reduce hallucinations. 

\subsection{No-Thinking/Thinking Mode Prompt}
\promptboxfromfile{No-Thinking Mode}{prompt/no-think-prompt.txt}
\promptboxfromfile{Thinking Mode}{prompt/think-prompt.txt}


\end{document}